\documentclass[journal]{IEEEtran}

\usepackage{times}
\usepackage{epsfig}
\usepackage{graphicx}
\usepackage{amsmath}
\usepackage{amssymb}
\usepackage{url}
\usepackage{txfonts}
\usepackage{microtype} 

\def\labs{\mathbb{N}}
\def\real{\mathbb{R}}

\long\def\comment#1{}

\newtheorem{mydef}{Definition}

\usepackage{pgfplots}

\usepackage{caption}
\usepackage{cite} 
\usepackage[pagebackref=true,breaklinks=true,letterpaper=true,colorlinks,bookmarks=false]{hyperref}

\newcommand{\yearplot}[2][]{%
\hypersetup{citecolor=black}
\begin{tikzpicture}
\begin{axis}[
  xlabel=Year,
  x tick label style={/pgf/number format/1000 sep={}}, 
  ylabel style={align=center}, 
  legend entries={Metric, Nonmetric, Nonmetric w/ Side-info},
  legend pos=south east,
  xtick={2000,...,2050},
  minor x tick num=1, 
  xtick align=outside,
  xminorgrids=true,
  ymajorgrids=true,
  minor grid style={draw=lightgray, line width=1pt},
  minor tick style={draw=lightgray, line width=1pt},
  minor tick length=10pt,
  #1,
]
\addplot[
          only marks,
          scatter,
          scatter/classes={y={mark=triangle*, scale=1.3, draw=blue!70!gray!60!white, fill=white},
                           n={mark=square*,draw=red!70!black, fill=white},
                           s={mark=square*,draw=green!70!black, fill=white}},
          scatter src=explicit symbolic,
          mark size=0.2cm,
          line width=1pt,
          nodes near coords*={\textbf{\footnotesize\citen\bibref}},
          visualization depends on={value\thisrow{bib_ref} \as \bibref},
          nodes near coords align={},
]
table[x=year, y=accuracy,  y error=error, meta=metric]{#2};
\end{axis}
\end{tikzpicture}
\hypersetup{citecolor=green}
}

\begin{document}
\title{Good Recognition is Non-Metric}
\author{Walter J. Scheirer$^{1,3,4}$, Michael J. Wilber$^{1,3}$, Michael
  Eckmann$^2$, and Terrance E. Boult$^{1,3}$ \\
  $^1$ University of Colorado, Colorado Springs \\
  $^2$ Skidmore College \\
  $^3$ Securics, Inc \\
  $^4$ Harvard University \\
  \{\texttt{wscheirer,mwilber,tboult}\}\texttt{@securics.com}, ~\texttt{meckmann@skidmore.edu}
}
\maketitle

\begin{abstract}
Recognition is the fundamental task of visual cognition, yet how to formalize the general recognition problem for computer vision remains an open issue. The problem is sometimes reduced to the simplest case of recognizing matching pairs, often structured to allow for metric constraints. However, visual recognition is broader than just pair matching -- especially when we consider multi-class training data and large sets of features in a learning context. What we learn and how we learn it has important implications for effective algorithms. In this paper, we reconsider the assumption of recognition as a pair matching test, and introduce a new formal definition that captures the broader context of the problem. Through a meta-analysis and an experimental assessment of the top algorithms on popular data sets, we gain a sense of how often metric properties are violated by good recognition algorithms. By studying these violations, useful insights come to light: we make the case that locally metric algorithms should leverage outside information to solve the general recognition problem.
\end{abstract}

\section{Introduction}
\label{sec:intro}
Recognition is a term everyone in computer vision and machine learning understands --  or at least we think we do.  Despite multiple decades of research, it may be somewhat surprising to learn that a very basic question remains unresolved: \textit{is recognition metric}? Familiar distance metrics used in vision include Euclidean distance and Mahalanobis distance, both computed in feature space. Given one of these metrics, the task of recognizing an unknown object can be approached by finding the class label of its nearest neighbor under that distance metric in a set of training samples. Clearly such an approach provides a recognition function, so some level of recognition can be accomplished with a metric.  However, at a more fundamental level, we would like to know if distance truly captures all that is meant by the term recognition, and if metrics are good approaches to solving complex recognition tasks in computer vision. In this paper, we adopt the convention that a problem is metric if the best solutions to that problem can be achieved by directly applying a distance metric to compute the answer.

An important observation with implications for recognition is that in separable metric space, using a distance metric and the nearest neighbor (NN) algorithm provides useful consistency. As the number of i.i.d. samples from the classes approaches infinity, the NN algorithm will converge to an error rate no worse than twice the Bayes error rate, \textit{i.e.} no worse than twice the minimum achievable error rate given the distribution of the data~\cite{cover1967nearest}. To many, this convergence theorem suggests that recognition can always be formulated as NN matching with an appropriate distance metric. However, having to double the error of the optimal algorithm over the same data often does not lead to a particularly good algorithm. This becomes apparent when actual error rates are considered during experimentation.

With the recent popularity of metric learning~\cite{Yang_2006, Frome_2007, Kulis_2009, Cao_2010, nguyen-2010, ying-and-li-2012, Jain_2012} for various recognition tasks, where a metric is learned over given pairs of images that are similar or dissimilar, one might infer that recognition is always a metric process. We note that the NN convergence theorem~\cite{cover1967nearest} is true for {\em any} metric -- hence any improvements from the choice of metric, or metric learning, are not about the asymptotic error, but something else such as the error for finite samples and/or the rate of convergence. This paper will show that while metric learning can produce reasonable results, enforcing metric properties leaves out  information, often limiting the quality of recognition with finite data.

\begin{figure}
\centering
\includegraphics [width=0.9\linewidth]{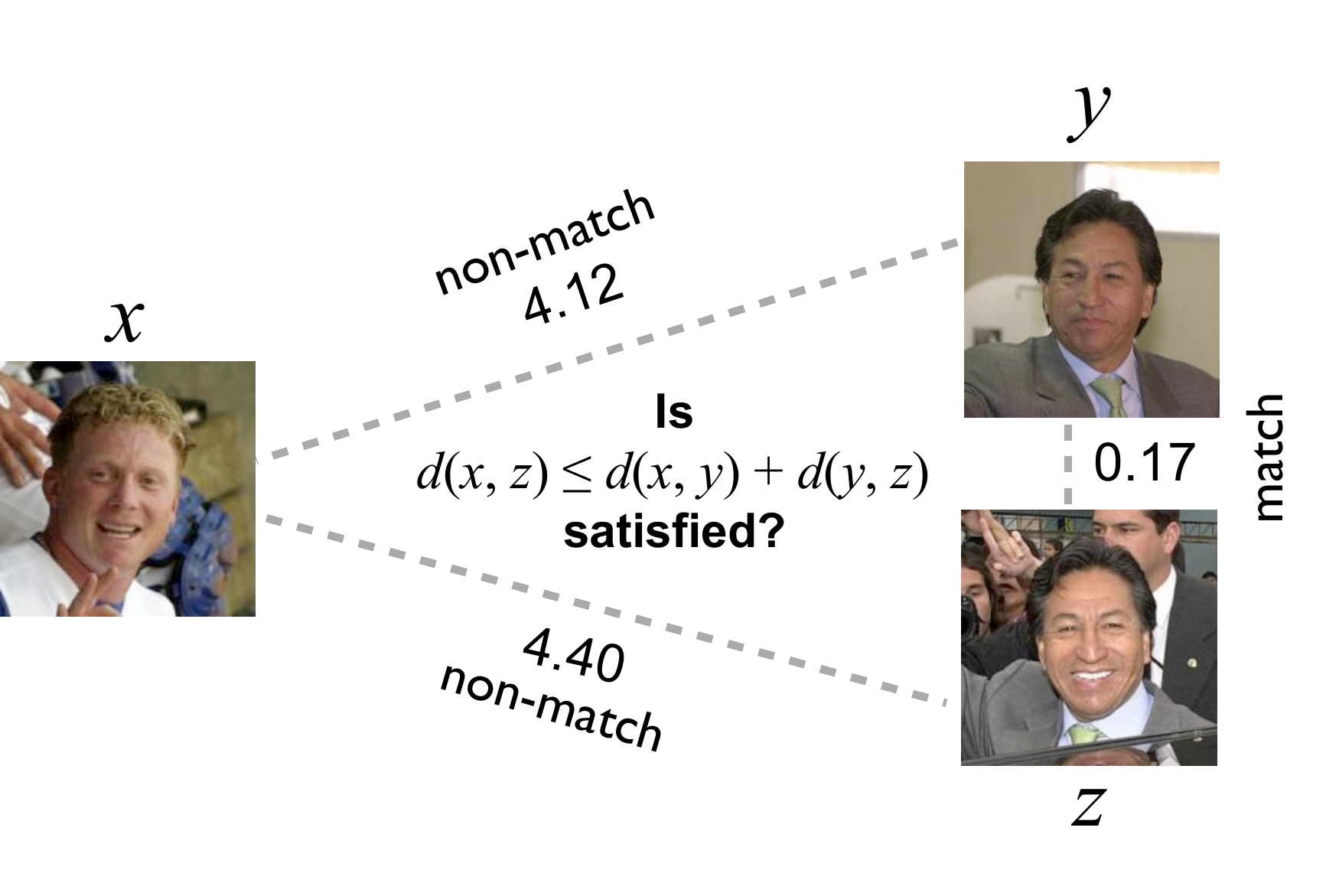}
\caption{Assumptions are often made about the underlying nature of recognition in computer vision that do not hold true in practice. A common constraint placed upon recognition algorithms is that they must be \textit{metric}, meaning their distance scores adhere to the properties of non-negativity, identity, symmetry and the triangle inequality. At first glance, the scores from many recognition algorithms appear to satisfy these constraints. However, violations can be subtle. For example, the distance scores produced by the top-performing Tom-vs-Pete algorithm~\cite{berg_2012} for these images from LFW~\cite{LFWTech} violate the triangle inequality.}
\label{fig:teaser}
\end{figure}

If the convergence theorem itself is about recognition, then the recognition problem is {\em assumed} to be formulated in an asymptotic sense with infinite i.i.d. samples.  We argue that visual recognition does not rely on either of those assumptions, but rather focuses on maximizing the accuracy for finite, and, unfortunately, opportunistic and hence potentially biased sampling.

Recall that a function $d: X \times X  \to \real$ is metric over a set $X$ if it satisfies four properties for $\{x,y,z\} \subseteq X$:
\begin{enumerate}
\itemsep 0pt
\parskip 0pt
\item $d(x, y) \ge 0$ (non-negativity)
\item $d(x, y) = 0 \Leftrightarrow x = y$ (identity)
\item $d(x, y) = d(y, x)$ (symmetry)
\item $ d(x, z) \le d(x, y) + d(y, z)$ (the triangle inequality)
\end{enumerate}
Metric functions have useful properties that allow one to show that a particular problem leads to a convex minimization problem, or that various types of sequences converge in the limit. There are also several cases where one of the properties is excluded. Functions that do not satisfy the triangle inequality are called semimetrics, those that violate symmetry are called quasimetrics, and those missing one or both halves of the identity requirement are called pseudometrics\footnote{Note that without the property of identity, the theorem of NN convergence~\cite{cover1967nearest} does not hold. It has also been shown~\cite{MahamudICCV2003} that the optimal distance measure, in the sense of minimal Bayes risk always violates the identity property and therefore is not metric.}. While the term ``distance measure'' is sometimes used to mean a distance metric, it is more appropriate to use this term to mean a measurement that provides information about dissimilarity, but may be formally non-metric (our use of the term follows this convention).

A natural place to begin examining if recognition is metric is to consider its formulation.
Is it reasonable to assume that a distance metric $d$ maps pairs of elements from $X$ into $\real$ during recognition? When a person recognizes an object, do they refer to an actual image of the object of interest?
A more likely alternative is a comparison to a stored model with a more complex internal representation, not a direct copy of some prior trained input. This view is consistent with prototype theory~\cite{rosch_1976} in cognitive psychology. Thus, at a structural level, recognition in this mode takes an input $x\in X$, and a model $M$, and hence cannot be metric because it is not even of the proper functional form. It is possible to build a model using just  $x$, and then consider the distance between models.  However, for many commonly used recognition algorithms, \textit{e.g.} support vector machines, one cannot induce a proper model from a single input. Thus, the general problem of recognition cannot be restricted to just metrics, even though it must include them.

To structure the recognition problem in a form consistent with directly using a distance metric, various approaches formulate it as a pair matching task. This is a specific case of recognition: given a pair of images, an algorithm must indicate whether or not they match (the popular face recognition challenge problem Labeled Faces in the Wild~\cite{LFWTech} is of this form). While it is convenient to build a binary classifier for this type of problem and easier to evaluate this type of data, pair matching itself cannot be considered the general recognition problem, where $n$ number of known classes might be candidates during matching, represented by complex models incorporating information from potentially thousands of training images.

It is also natural to ask if the human mind, a most successful recognition system, operates in a way that satisfies the key metric properties of symmetry and the triangle inequality.  The consensus in the cognitive psychology community is a definitive no. In seminal work, Tversky~\cite{Tversky_1977} showed that human analysis of ``similarity'' is non-symmetric and is context dependent. One of the visual experiments conducted by Tversky was a simple pair matching task, where subjects were asked if two block letters were the same or not. A similarity function ${\cal S}(p,q)$ indicated the frequency at which subjects noted letter $p$ to be the same as $q$.  The experiment showed that the order of presentation of the letters mattered in a statistically significant way:  ${\cal S}(p,q) \ne {\cal S}(q,p)$.
This result, along with others for matching faces, abstract symbols, and the names of countries led Tversky to conclude that ``similarity is not necessarily a symmetric relation.''

In subsequent work, Tversky and Gati~\cite{Tversky_1982} examined if the triangle inequality is satisfied by humans when assessing similarity. Because the triangle inequality can always be satisfied by adding a large constant to the distances between individual points when measuring dissimilarity on an ordinal scale, Tversky and Gati proposed a test that assumes segmental additivity: $d(x, z) = d(x, y) + d(y, z)$. Over numerous pair matching trials across stimuli, human similarity judgments were found to violate the triangle inequality in a statistically significant manner. Even without the triangle inequality for additive functions, it is still possible to induce metric models with subadditive metrics. However, in experiments where subjects provided subjective probability estimates, Tversky and Koehler~\cite{tversky1994support} showed that the reported scores are, in general, not subaddative\footnote{It is possible to work around the constraint of segmental additivity using a subadditive metric based on Shepard's universal law of generalization to induce a metric from finite sets of data~\cite{jakel_2008}, but the result is still not consistent with the human perception findings of Tversky and Koehler~\cite{tversky1994support}. }.

{\em If humans are employing non-metric, non-symmetric similarity measures, do we really want to constrain our recognition algorithms in computer vision to be metric?} Addressing this notion, we present the following contributions:
\begin{itemize}
\item A new general definition of recognition that is not restricted to pair matching, and which includes provisions for complex models trained over sets of images and assumptions.
\item An extensive meta-analysis of metric learning on vision problems, along with experiments that give an indication of how often metric constraints are violated for top performing algorithms and common data sets.
\item A series of useful recommendations, based on our results, for recognition algorithm designs in metric and non-metric spaces.
\end{itemize}

\section{A General Definition of Recognition}
\label{sec:definition}

Surprisingly, a canonical definition of recognition for computer vision has yet to emerge. Many different definitions of recognition can be found in the literature, each addressing particular aspects of the problem. The familiar distance-based approach to recognition~\cite{Frome_2007,Kulis_2009,Jain_2012} compares feature vectors from a test image to one or more feature vectors from known images using a distance measure to indicate similarity. More compatible with recent machine learning-based approaches, statistical learning theory~\cite{vapnik-82} casts recognition as risk minimization over a given loss function and joint probability distribution for a class. Other definitions include the probabilistic formulation described by Shakhnarovich et al.~\cite{darrell-etal-eccv02}, where recognition maximizes the probability that an input distribution matches a probability rule for a single known class, as well as the ubiquitous nearest neighbor decision rule~\cite{cover1967nearest} discussed above.

With many possibilities for class sampling, modeling for training, and strategies for matching, a concise definition that captures all of these aspects is an open issue. The above definitions tend to satisfy the definition of a particular subproblem in recognition, such as pair matching (1:1 matching), verification (1:1 matching with any claimed class), identification (1:$n$ matching), or search (1:$n$ matching returning multiple results). However, no current definition captures the general problem encompassing all of them. Further, each definition is missing necessary detail with respect to the information available during matching. For a given class, there is a possibility that assumptions outside any given training examples have been made, which should be incorporated into the overall definition. These assumptions can include side-information~\cite{Xing_2002}, regularization terms~\cite{learning-with-kernels-02}, score normalization~\cite{scheirer_cvpr2012}, or more fundamentally, data used to train a detector that is applied when pre-processing the training and testing images. Another consideration is the possibility of nested or hierarchical classes, where it is necessary to return multiple class labels for a given input. With all of these issues in mind, we introduce the following more comprehensive definition:

\begin{mydef}
\setlength\abovedisplayskip{0pt}
\setlength\belowdisplayskip{2pt}

\textup{(The General Recognition Problem)}
Given image(s) $I \in \real^\nu$, where $\nu$ is the number of pixels,
let $F:\real^\nu \to \real^D$ extract a $D$-dimensional feature vector $x$ under a set of feature extractor-specific assumptions $\phi_F$:\\
\vspace*{-1ex}
 \begin{equation}
x = F(I,\phi_F), x \in \real^D
\label{eq:feature}
\end{equation}
The task of a recognition system is to find a ranked set of integer class labels considered to be the best matches to a given input feature vector $x_0$. For a class labeled $c\in\labs$, let $X_c$ be a set of training data $\{x_1, \ldots\}$ composed of $m$ feature vectors, where $m \ge 1$. A class model $M_c$ represents the information learned from $X_c$, incorporating a set of modeling-specific assumptions $\phi_M$. Let $R$ be a matching function that produces a similarity score $s$ by comparing $x_0$ to $M_c$, taking into account a set of matching-specific assumptions $\phi_R$:\\
\vspace*{-1ex}
\begin{equation}
s_c = R(x_0, M_c(X_c, \phi_M), \phi_R), s_c \in \real
\label{eq:score}
\end{equation}
For any input $x_0$, let $S$ be a set of similarity scores $\{s_1, \ldots\}$ from $n$ matching instances of $R$, where $n \ge 1$. Let $L$ be a labeling function that maps $S$ to a ranked set of $k$ class labels $C = \{c_{1}^*, \ldots\}$, where $k \ge 1$, taking into account any labeling-specific assumptions $\phi_L$:\\
\vspace*{-1ex}
\begin{equation}
C = L(S, \phi_L), C \subsetneq \labs
\label{eq:label}
\end{equation}
where $c_1^*=0$ is reserved for the non-match label.

\label{def:recognition}
\end{mydef}

Def.~\ref{def:recognition} is consistent with the four modes of recognition described above:
\begin{enumerate}
\itemsep 1pt
\parskip 0pt
\item For pair matching, $M_c$ can consist of just features from a single training image $X_c = x_1$, with $R$ a distance measure between vectors and $|C| = 1, c^* \in \{0,1\}$ (non-match and match). $\phi_L$ contains matching criteria (\textit{e.g.} an estimated threshold). $M_c$ can also be a complex model over many images, matching against the image pair as $x_0$ (see the discussion of LFW in Sec.~\ref{sec:lfw-meta-analysis}).
\item For verification,  we seek to check if an input image belongs to a class $c$ specified \textit{a priori}, with training data defined as above for pair matching. $R$ could be applied $n$ times in a multi-view setting with multiple models, matching against the set $\{M_{c_{1}}, \ldots\}$ for class $c$,  where $n \ge 1$. In all cases, $\forall c^* \in C, c^* \in \{0,c\}$ and $\phi_L$ contains matching criteria.
\item Identification can also make use of the same training strategies as pair matching, but always applies $R$ over a set of $n$ different classes, where $n \ge 2$. It returns at most one best answer with $k=1$.
\item Search is similar to identification, but returns multiple labels, \textit{i.e.} $k>1$.
\end{enumerate}

\comment{. For nearest neighbor classification~\cite{cover1967nearest}, each class model $M_c$ consists of positive training data $X_c = \{x_{1}^+,\ldots, x_{m}^+\}, m \ge 1$, $R$ is a distance function, and $L$ assigns $c_1^*$ (the top match) by finding $\inf_c S$. For support vector machine (SVM) classification~\cite{learning-with-kernels-02}, assumptions $\phi$ should include any free parameters (regularization term, bias term, kernel parameters). A model $M_c$ is trained over positive and negative data $X_c = \{x_{1}^{+},\ldots, x_{m}^+,x_{1}^{-},\ldots, x_{m}^-\}, m \ge 1$, $R$ is a linear classifier or a non-linear kernel machine producing similarity scores $S$ that represent distance from the margin, and $L$ assigns $c^* \in \{0,1\}$ in the binary case, and $c^* \in \{0, \ldots, n\}$, where $n \ge 2$, in the multi-class case. For one-shot learning~\cite{feifei_2006}, a Bayesian model $M_c$ can be trained with a single example $x_{1}^+$ from the positive class and $m$ priors $x^{p}_i$ over data from other classes: $X_c = \{x_1, x_{1}^p \ldots, x_{m}^p\}$. In this case, $R$ produces a probability, and $L$ assigns $c^* \in \{0,1\}$.}

\section{Meta-Analysis of Algs. for LFW}
\label{sec:lfw-meta-analysis}

Our first case study is Labeled Faces in the Wild~\cite{LFWTech}, a
popular data set among face recognition researchers.
LFW is ideal for testing pair matching algorithms because it is
inherently a pair matching problem. Using the terminology of Def.~\ref{def:recognition}, each algorithm selects an appropriate feature representation
$F$, a model representation $M_c$, and a matching function $R$. Each input
is a pair of feature vectors. For consistency
with Def.~\ref{def:recognition}, we express this as the
concatenation of the two fixed-length input vectors; thus, $x_0 =
F(I_{1}, \phi_F) \, \| \, F(I_{2}, \phi_F)$ where $\|$ denotes
concatenation. Likewise, each algorithm may train on $X = \left\{
  x_1^+, \ldots, x^+_{m}, x_{m+1}^-, \ldots x^-_{2m}\right\}$, a set
of $m$ matching pairs and $m$ nonmatching pairs of features. The
labeling function $L(S, \phi_L)$ usually checks some likelihood against a threshold $\tau$ to
decide whether the pair matches, returning $c^*=1$ if $s_1 > \tau$
and $c^*=0$ otherwise, but certain algorithms may instead define
something more complicated.

\begin{figure*}[hbtp]
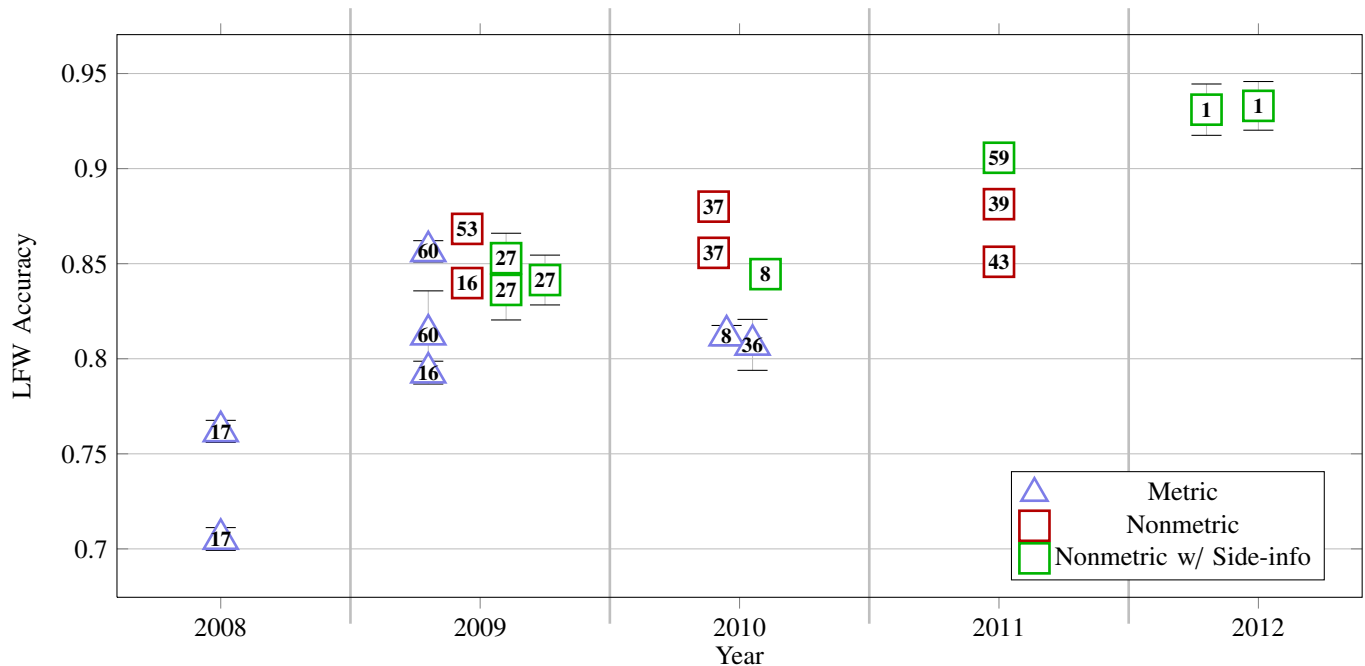

\yearplot[width=\textwidth,
          height=0.5\textwidth,
          error bars/y dir=both,
          error bars/y explicit,
          error bars/error bar style={draw=lightgray},
          ylabel={LFW Accuracy}]{lfw-table.dat}
          \caption{\label{fig:meta-analysis-lfw}Recognition accuracy
            of algorithms on LFW. Horizontal axis is year of
            publication; some cluttered years are slightly separated
            along the horizontal axis for clarity. ``Side-info''
            refers to algorithms that use outside data in the
            recognition system beyond feature extraction/alignment.
            Even for ``pair matching,'' pure metric algorithms are not
            very competitive. Numbers inside each point correspond to
            bibliography entries. A table of raw scores is shown in Table~\ref{tab:lfw}}
\end{figure*}

In this analysis, we consider only recent results for the
``Image-restricted'' setting where outside data was used for feature
extraction and in the recognition system. We chose this set of results
because it represents several algorithms that are both metric and
non-metric, allowing us to compare the performance of both.
To avoid confirmation bias, we only investigate the 20
results listed on the official LFW results web page at the time of
writing~\cite{lfw-results}. How well do metric learning algorithms perform on this slice of LFW?
By graphing the accuracy of these results over time, some
interesting trends become apparent; see Fig.~\ref{fig:meta-analysis-lfw}.

First, with the exception of~\cite{ying-and-li-2012}, \emph{the
  non-metric algorithms perform better} than the algorithms that
constrain themselves to be completely metric.
We investigate specific cases below.
Second, \emph{the first results reported on LFW are from metric
  learning algorithms}, but more recent results are not metric and do
not claim to be metric.
Note that in Fig.~\ref{fig:meta-analysis-lfw}, we only consider an algorithm
to be ``metric'' if the scoring function $R$ completely satisfies all
four properties of distance metrics outlined in
Sec.~\ref{sec:intro}.
Though many of the
papers claim to be metric, upon closer investigation, some of them
have a non-metric $R$ or only use metric learning as a part of their
overall computation. For example, some techniques define an $R$ that
combines locally metric information over different neighborhoods,
making $R$ globally non-metric.

One example of an algorithm that turns out to be non-metric is~\cite{guillaumin-2009}, which
uses a custom logistic discriminant-based metric learning (LDML)
approach. The algorithm specifies a nearest-neighbor-like (MkNN)
normalization strategy: during testing time, each pair's score is
influenced by neighborhoods of matching pairs around the two images
being compared. In our words, they define a recognition function
$R_{\text{MkNN}}(x_0, M_c(X_c, \phi_M(x_0)), \phi_R)$. Note that
$M_c(X_c, \phi_M(x_0))$ now changes at test time: instead of
$R_{\text{MkNN}}$ being fixed on a particular global model, each
model's assumptions $\phi_M(x_0)$ depend on the input testing pair.
From this, it is easy to see that LDML-MkNN is not globally
metric: $R_{\text{MkNN}}$ no longer satisfies the triangle inequality
because it depends on a model with assumptions that change as a function of the
image pair being classified.

\hypersetup{citecolor=black}
\begin{table}[h]
  \centering
  \small
\begin{tabular}{lrlrr}
Year & Accuracy&& Metric? & Citation\\
\hline
2012 & 93.30\% & $\pm$ 1.28\% & No, w/ sideinfo & \cite{berg_2012} \\
2012 & 93.10\% & $\pm$ 1.35\% & No, w/ sideinfo & \cite{berg_2012} \\
2011 & 90.57\% & $\pm$ 0.56\% & No, w/ sideinfo & \cite{yin_et_al_2011} \\
2011 & 88.13\% & $\pm$ 0.58\% & No & \cite{pinto_2011} \\
2010 & 88.00\% & $\pm$ 0.37\% & No & \cite{nguyen-2010} \\
2009 & 86.83\% & $\pm$ 0.34\% & No & \cite{wolf_2009} \\
2009 & 85.65\% & $\pm$ 0.56\% & Yes & \cite{ying-and-li-2012} \\
2010 & 85.57\% & $\pm$ 0.52\% & No & \cite{nguyen-2010} \\
2009 & 85.29\% & $\pm$ 1.31\% & No, w/ sideinfo & \cite{kumar_et_al_2009} \\
2011 & 85.10\% & $\pm$ 0.59\% & No & \cite{seo_2011} \\
2010 & 84.45\% & $\pm$ 0.46\% & No, w/ sideinfo & \cite{Cao_2010} \\
2009 & 84.14\% & $\pm$ 1.31\% & No, w/ sideinfo & \cite{kumar_et_al_2009} \\
2009 & 83.98\% & $\pm$ 0.35\% & No & \cite{guillaumin-2009} \\
2009 & 83.62\% & $\pm$ 1.58\% & No, w/ sideinfo & \cite{kumar_et_al_2009} \\
2009 & 81.27\% & $\pm$ 2.30\% & Yes & \cite{ying-and-li-2012} \\
2010 & 81.22\% & $\pm$ 0.53\% & Yes & \cite{Cao_2010} \\
2010 & 80.73\% & $\pm$ 1.34\% & Yes & \cite{nair_2010} \\
2009 & 79.27\% & $\pm$ 0.60\% & Yes & \cite{guillaumin-2009} \\
2008 & 76.18\% & $\pm$ 0.58\% & Yes & \cite{Huang_2008} \\
2008 & 70.52\% & $\pm$ 0.60\% & Yes & \cite{Huang_2008} \\
\end{tabular}
\caption{\label{tab:lfw} A table of accuracy from LFW results}
\end{table}
\hypersetup{citecolor=green}

Even without the MkNN step, we can make the case
that the implementation of LDML is non-metric.
According to Sec.~2 of~\cite{guillaumin-2009}, the $R$ defined by
the base algorithm is
$R(x_0, M_c, \phi_R) = \sigma(b - d_W(F(I_1,\phi_F), F(I_2,\phi_F)))$,
where $b$ is a bias term, $\sigma$ is the sigmoid function, and $d_W$
is the Mahalanobis-like measure. Rather than actual covariance, $W \in
\real^{D\times D}$ is a learned matrix, part of model $M_c$. If $W$
was symmetric and positive-definite, it would result in a metric.
However, in Sec.~2.3 of~\cite{guillaumin-2009}, it is stated that
no such constraints are placed on $W$. Thus, this learned distance may
not be even pseudometric.

Another example of an algorithm that turns out to be non-metric is Cosine Similarity Metric
Learning as presented in~\cite{nguyen-2010}.
According to Sec. 1.2 of~\cite{nguyen-2010}, $R$ is
defined as
\begin{equation}
\vspace*{1ex}
  R_{\text{CSML}}(x_0, M_c(X_c, \phi_M), \phi_R) = \frac{(a_n)^T (b_n)}{ \|a_n\| \; \|b_n\|} = \cos \theta
\end{equation}
where $a_n$ and $b_n$ are $A(F(I_{n,1}, \phi_F))$ and $A(F(I_{n,2},
\phi_F))$ for some matrix $A$, part of model $M_c$ that is learned to
minimize the distance between positive pairs and maximize the distance
between negative pairs. The algorithm's labeling assumption $\phi_L$ is a threshold $\tau$ over
$\cos\theta$, where $\theta$ is the angle between $a_n$
and $b_n$. However, $\cos$ is not a distance metric since it only
satisfies one of the four properties outlined in
Sec.~\ref{sec:intro}. First, note that $\cos$ is bounded by -1 and
1, but distances must not be negative. Second, $\cos$ may be 0 if
$a_n$ and $b_n$ are perpendicular, so $s_c=0$ does not imply that
$a_n$ and $b_n$ are the exact same input. This also means $\cos$ does not
satisfy the triangle inequality. The only metric property that CSML
satisfies is symmetry.

A significant advantage of CSML is that the boundedness of $R_{\text{CSML}}$
between -1 and 1 allows for a fast coarse-to-fine search for optimal parameters. In
fact, many algorithms use metric learning precisely for this reason.
Here, CSML has found one way to use this property while still
performing better than other learning techniques, even though it is
not actually metric.

Another system that incorporates metric learning as part of a
pipeline that is not completely metric is~\cite{Taigman_2009}, which
uses multiple one-shot similarity (OSS). In standard OSS, two models
are trained at test time from canonical ``negative'' examples with
each image in the image pair as positives:
\begin{equation}
\begin{split}
\hat{M_c}(X_c, \phi_M) = \{M_{1}(F(I^+_1, \phi_F), x_1^-, \ldots), \\
M_{2}(F(I^+_2, \phi_F), x_1^-, \ldots)\}
\end{split}
\end{equation}
Then, the scoring function $R_{\text{OSS}}(x_0,
\hat{M_c}(X_c, \phi_M), \phi_R)$ uses each model to classify its
respective input and averages the two scores. However, when labeled
information is available, there is no clear way for OSS to take
advantage, and thus OSS may be biased toward pose, lighting, etc. To
get around this, another algorithm Multi-OSS is defined, which computes multiple
one-shot scores for multiple labels at test time, increasing the
generality of the classifier. Note that neither OSS nor Multi-OSS are
metric because each score depends on models created at testing time,
each using different assumptions/examples. However,~\cite{Taigman_2009}
shows that OSS and Multi-OSS are more effective than a variety of metric techniques.
The improvement is attributed
to the extra information provided by the class labels -- something that
the metric techniques cannot take advantage of.

According to Fig.~\ref{fig:meta-analysis-lfw}, we see that the top
scores come from non-metric algorithms, whether the authors intended
them to be metric or not. What makes non-metric algorithms better?
We emphasize that treating all samples alike may unnecessarily
handicap an algorithm. For
example, if one classifier is more invariant to pose, that classifier
may be better than a generic classifier at handling samples with
differing pose. This approach is embraced in~\cite{Cao_2010},
where
several SVM classifiers are trained across different subsets of the gallery
for each pose combination to create a pose-adaptive classification system.
Similarly, the top performing algorithm on the LFW unrestricted set ~\cite{Li_2012} uses a
probabilistic model based on the observation that features extracted
from an image can change with respect to irrelevant variables such as
pose, expression, and illumination, which may dwarf the variation
created by the actual change in identity in the image pair~\cite{Li_2012}.  A perfect metric system
must filter out such unwanted variation completely, which is
impossible if all variables can influence score distances.

\begin{figure*}[t]
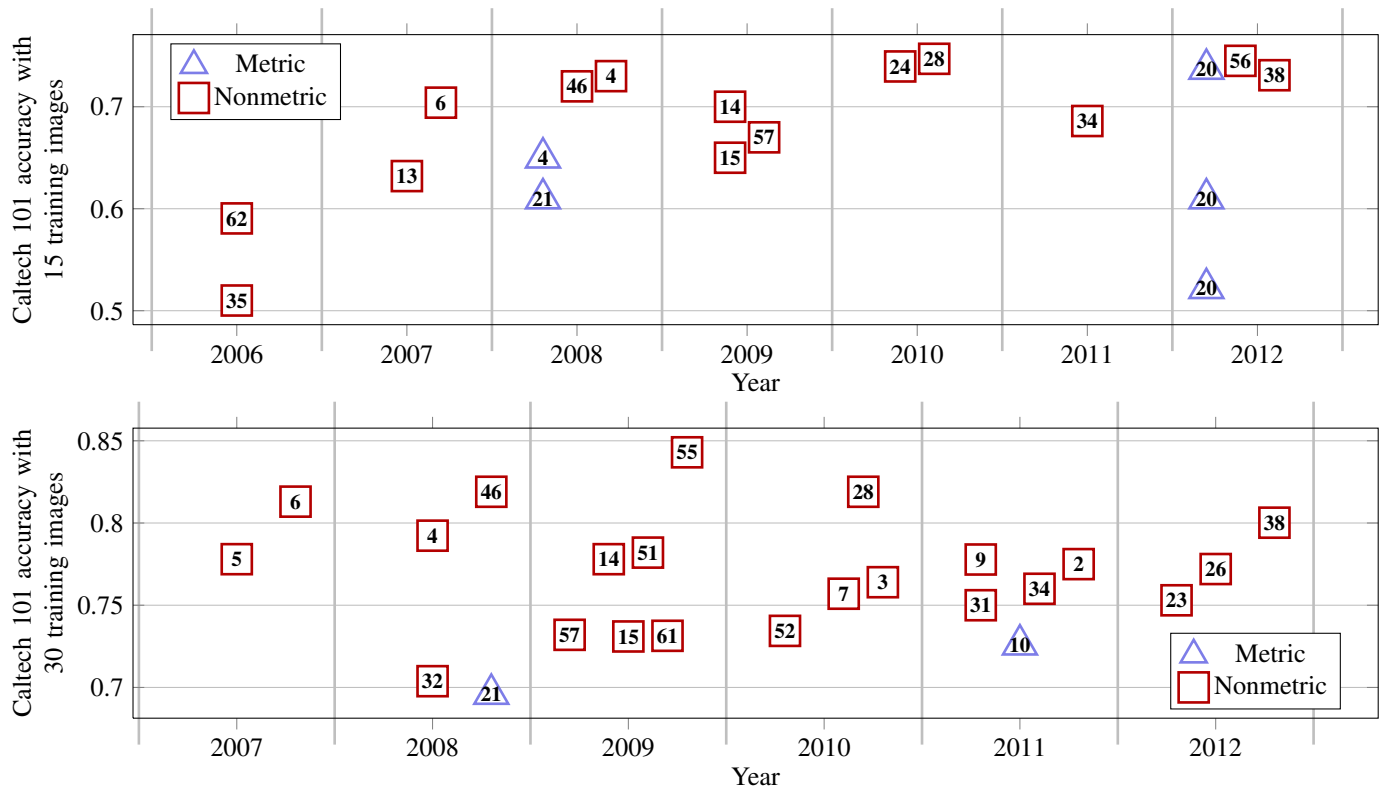

\yearplot[width=\textwidth,
          height=0.3\textwidth,
          legend entries={Metric, Nonmetric},
          legend pos={north west},
          ylabel={Caltech 101 accuracy with\\15 training images}]{caltech-table15.dat}
\yearplot[width=\textwidth,
          height=0.3\textwidth,
          legend entries={Metric, Nonmetric},
          ylabel={Caltech 101 accuracy with\\30 training images}]{caltech-table30.dat}
          \vspace{-2ex}
          \caption{\label{fig:meta-analysis-ct101}Recognition accuracy
            of algorithms on Caltech 101, with 15 training images on
            the top plot and 30 on the bottom plot. The horizontal
            axis is year of publication; some cluttered years are
            slightly separated along the horizontal axis for clarity.
            Note the metric algorithms are generally not as accurate,
            but are more competitive when fewer images can be used for
            training. Numbers inside each point correspond to
            bibliography entries. Note that because not all algorithms
            reported error bars, we do not show any error bars in this
            plot. A table of raw scores is available in
            Table~\ref{tab:caltech15} for 15 samples and
            Table~\ref{tab:caltech30} for 30 samples.
        }
\end{figure*}

\hypersetup{citecolor=black}
\begin{table}[h]
  \centering
  \small
\begin{tabular}{lrlrr}
Year & Accuracy&& Metric? & Citation\\
\hline
2010 & 74.70\% &  & No & \cite{li2010object} \\
2012 & 74.50\% &  & No & \cite{yang2012} \\
2010 & 73.95\% & $\pm$ 1.13\% & No & \cite{kapoor2010gaussian} \\
2012 & 73.70\% &  & Yes & \cite{Jain_2012} \\
2012 & 73.09\% &  & No & \cite{oliveira2012sparse} \\
2008 & 73.00\% &  & No & \cite{boiman2008defense} \\
2008 & 72.00\% &  & No & \cite{Todorovic_learningsubcategory} \\
2007 & 70.40\% & $\pm$ 0.70\% & No & \cite{bosch2007image} \\
2009 & 70.00\% &  & No & \cite{gehler2009feature} \\
2011 & 68.60\% &  & No & \cite{mccann2012local} \\
2009 & 67.00\% &  & No & \cite{yang2009linear} \\
2009 & 65.00\% &  & No & \cite{gu2009recognition} \\
2008 & 65.00\% &  & Yes & \cite{boiman2008defense} \\
2007 & 63.20\% &  & No & \cite{Frome_2007} \\
2008 & 61.00\% &  & Yes & \cite{jain2008fast} \\
2012 & 61.00\% &  & Yes & \cite{Jain_2012} \\
2006 & 59.05\% & $\pm$ 0.56\% & No & \cite{zhang2006svm} \\
2012 & 52.20\% &  & Yes & \cite{Jain_2012} \\
2006 & 51.00\% &  & No & \cite{mutch2006multiclass} \\
\end{tabular}
\caption{\label{tab:caltech15} A table of accuracy of results on
  Caltech 101, for 15 training samples.}
\end{table}
\hypersetup{citecolor=green}

\section{Meta-Analysis of Algs. for Caltech 101}

Our second case study examines the Caltech 101 data set~\cite{caltech101}. Whereas LFW is ideal for analyzing pair matching algorithms, Caltech 101 is the most well known object recognition set for identification and search scenarios, making it a useful subject of study for these other classes of recognition.  We performed a meta-analysis of the top performing algorithms using the most possible training samples (30), as well as those that used 15 training samples.  To avoid confirmation bias, we report on the results of work organized by Lim~\cite{caltech101-30-results} for the 30 training samples as well as the algorithms compared in Yang et al.~\cite{yang2012} and Jain et al.~\cite{Jain_2012}.  We also report results for the top performing algorithms utilizing 15 training samples, as listed by Lim~\cite{caltech101-30-results}. Like our analysis of LFW, we can draw some interesting conclusions by considering the plots in Fig.~\ref{fig:meta-analysis-ct101}.

Notably, there is a general absence of metric methods in Fig.~\ref{fig:meta-analysis-ct101}.  For the algorithms making use of 30 training samples, only~\cite{jain2008fast} \&~\cite{coates2011importance} are metric---and they rank 23rd and 25th on the top results~\cite{caltech101-30-results}. The top 22 algorithms for 30 training samples are non-metric.  For 15 training samples, although several non-metric algorithms~\cite{yang2012, kapoor2010gaussian, li2010object} do outperform it, the technique of Jain et al.~\cite{Jain_2012} is metric and performs well. Specifically, Eq. 6 in~\cite{Jain_2012} is the matching function that corresponds to $R$ in Def.~\ref{def:recognition}, which is metric when the chosen kernel function $\kappaup_0(x,y)$ is metric. However, the lack of metric approaches with larger amounts of training data suggests that good performance is achieved by exploiting relationships beyond pairs of samples.  A common strategy for Caltech 101 is to learn a model for \textit{multiple classes} (often using an SVM with a non-metric kernel) in a 1-vs-All configuration, which violates the constraint of symmetry.

Analyzing a couple of specific cases that approach the problem from a metric perspective, we again find clear violations of metric assumptions. Instead of learning a global distance metric, the technique of Frome et al.~\cite{Frome_2007} learns a local distance measure for every feature vector in $X_c$ for all classes $c$ (resulting in a set of assumptions $\phi_M(X_1, \ldots, X_m)$ that help build $M_c$) using sets of image triplets incorporating a reference image, matching image, and non-matching image.
This approach is clearly non-metric because it intentionally maintains asymmetry; Sec. 3 of~\cite{Frome_2007} states ``Let $f_{j,m}$ be the $m$th feature vector from image $j$.  We assume a basic asymmetric distance from a single feature vector $f_{j,m}$ from one image to the set of features $F_i$ from another.'' The asymmetry is inherent in computing distance within image triplets that are specific to each reference image  $f_{j,m}$.

As another example, Yang et al.~\cite{yang2012} refer to kernel metrics throughout their article and while they do use kernel metrics to build models, the overall recognition system is non-metric at a structural level.
Like the algorithm of Frome et al.~\cite{Frome_2007}, this approach makes use of data dependent local models of groups, as opposed to global models over all of the training data. Relating this back to Def.~\ref{def:recognition}, $R$ includes group-sensitive kernel weights $\beta_i^{g}$ (Sec. IV.A.3 of~\cite{yang2012}) as part of its matching-specific assumptions $\phi_R(g) = \{\beta_i^{g}, \ldots, \beta_M^{g}\}$, where $M$ is the total number of kernels, and $g$ is a specific group.  Asymmetry is again inherent in this formulation -- by changing the selected group $g$, there is no guarantee that different weights will yield the same classification result.

\comment{To understand why this is so, metrics are only possible when the comparable quantities are of the same $type$, \textit{e.g.} both are images or both are feature vectors. For each image they extract a feature representation $F$ based on DCSIFT, DSIFT, SS \& PHOG, and use a kernel metric to create a model $M_c$ of each class.  To see that the overall system is non-metric, in their matching function $R$ they compare features $F$'s to complex models $M_c$'s which is inherently a non-metric.  For example, in their equations (5) (9) and (11) $x$ is our $F$ and $\beta_m$ is our $M_c$. They are comparing a point to a model which breaks the symmetry property of metrics.}

\hypersetup{citecolor=black}
\begin{table}[h]
  \centering
  \small
\begin{tabular}{lrlrr}
Year & Accuracy&& Metric? & Citation\\
\hline
2009 & 84.30\% &  & No & \cite{yang2009group} \\
2010 & 81.90\% &  & No & \cite{li2010object} \\
2008 & 81.90\% &  & No & \cite{Todorovic_learningsubcategory} \\
2007 & 81.30\% & $\pm$ 0.80\% & No & \cite{bosch2007image} \\
2012 & 80.02\% & $\pm$ 0.36\% & No & \cite{oliveira2012sparse} \\
2008 & 79.23\% &  & No & \cite{boiman2008defense} \\
2009 & 78.20\% & $\pm$ 0.40\% & No & \cite{vedaldi2009multiple} \\
2007 & 77.80\% & $\pm$ 0.80\% & No & \cite{bosch2007representing} \\
2009 & 77.80\% & $\pm$ 0.40\% & No & \cite{gehler2009feature} \\
2011 & 77.78\% & $\pm$ 0.56\% & No & \cite{chatfield2011devil} \\
2011 & 77.50\% &  & No & \cite{bo2011object} \\
2012 & 77.20\% &  & No & \cite{kumar2012binary} \\
2010 & 76.40\% & $\pm$ 0.70\% & No & \cite{bo2010kernel} \\
2011 & 76.00\% & $\pm$ 0.90\% & No & \cite{mccann2012local} \\
2010 & 75.70\% & $\pm$ 0.90\% & No & \cite{boureau2010learning} \\
2012 & 75.30\% & $\pm$ 0.70\% & No & \cite{jia2012beyond} \\
2011 & 75.00\% & $\pm$ 0.80\% & No & \cite{lu2011image} \\
2010 & 73.44\% &  & No & \cite{wang2010locality} \\
2009 & 73.20\% &  & No & \cite{yang2009linear} \\
2009 & 73.14\% &  & No & \cite{yu2009high} \\
2009 & 73.10\% &  & No & \cite{gu2009recognition} \\
2011 & 72.60\% &  & Yes & \cite{coates2011importance} \\
2008 & 70.38\% &  & No & \cite{ma2008learning} \\
2008 & 69.60\% &  & Yes & \cite{jain2008fast} \\
\end{tabular}
\caption{\label{tab:caltech30} A table of accuracy of results on
  Caltech 101, for 30 training samples.}
\end{table}
\hypersetup{citecolor=green}

\section{Experimental Results}
\label{sec:experiments}

\begin{figure}
\centering
\includegraphics [width=1.0\linewidth]{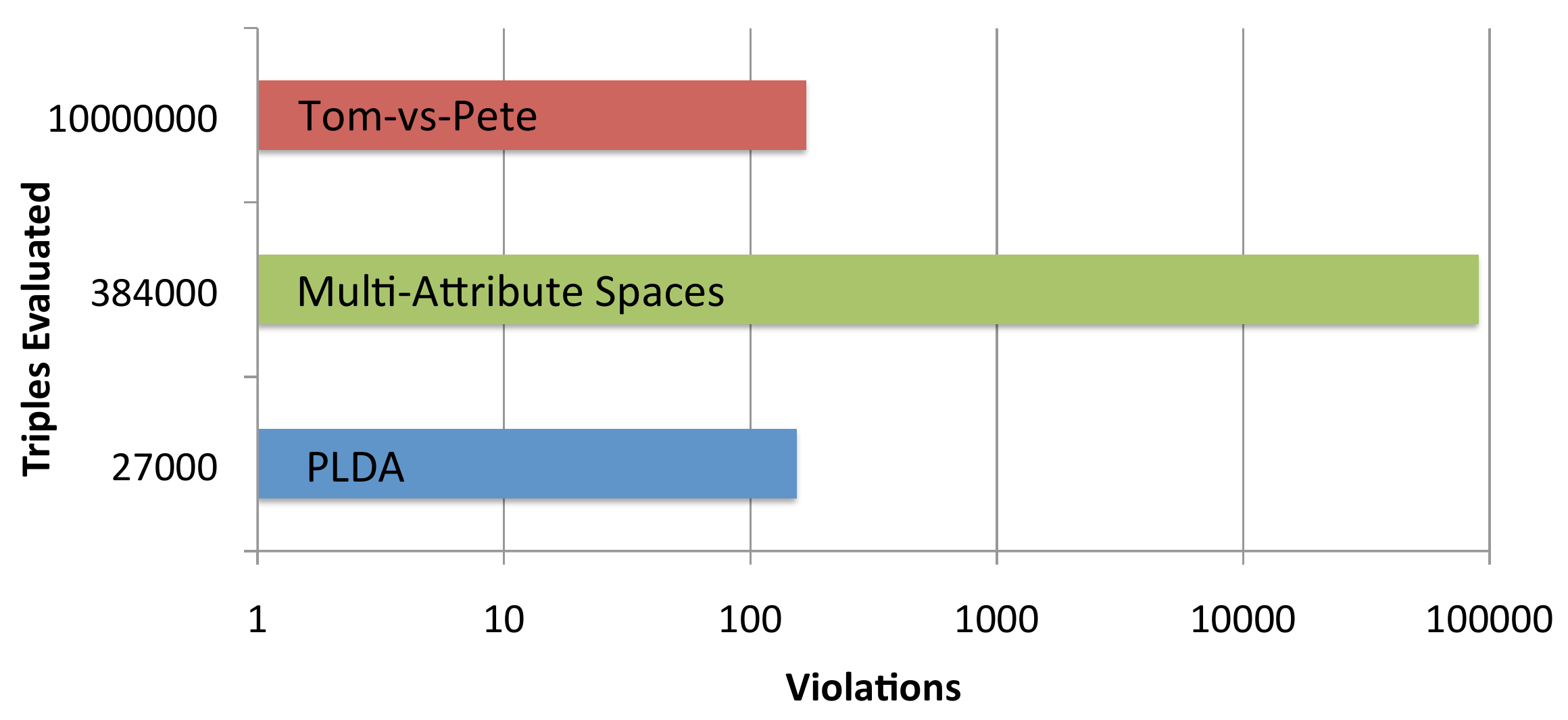}
\caption{Results showing the violations of the triangle inequality for three recent face recognition algorithms~\cite{berg_2012, scheirer_cvpr2012, Li_2012} applied over triplets of images drawn from the LFW~\cite{LFWTech} data set. Note that in some cases, it does not take a large sampling of triplets to find violations (PLDA), while in other cases, the occurrences are rare (Tom-vs-Pete), requiring a much larger evaluation.}
\label{fig:triangle}
\end{figure}

To gain a sense of how often the metric conditions are violated by good algorithms on pair matching tasks that appear to be metric in form, we conducted a series of experiments. Here we consider three different algorithms applied to data from LFW. The first algorithm is the ``Tom-vs-Pete'' classification approach of Berg and Belhumeur~\cite{berg_2012}, which learns a large set of identity classifiers, each trained over images for just two people. As of this writing, the ``Tom-vs-Pete'' algorithm is the best algorithm on the LFW Image-Restricted Training protocol. The second algorithm is the ``Multi-Attribute Spaces'' approach of Scheirer et al.~\cite{scheirer_cvpr2012}, where the statistical extreme value theory is leveraged to normalize scores across large sets of attribute classifiers for recognition tasks. The third algorithm is the ``Probabilistic LDA'' approach of Li et al.~\cite{Li_2012}, which uses a probabilistic generative model to determine if two faces have the same underlying identity cause. It is the best algorithm on the LFW Unrestricted Training protocol~\cite{lfw-results}.

Violations of the triangle inequality are subtle, requiring us to perform a large-scale search of the LFW image space. Triplets of images, such as the one shown in Fig.~\ref{fig:teaser}, are generated by sampling image combinations from the LFW set, including cases where matches and non-matches occur. Using each algorithm, we calculated the match score for each unique image pair in the triplet, and then checked if the scores satisfied the triangle inequality. To ensure a proper evaluation of distance, the scores $s_1, \ldots, s_n$ from the algorithms are processed with a simple transform $T$ that forces a ``smaller is better'' result: $T(s_i) = s_{\ell} - s_i$, where $s_\ell$ is the largest score in the set $\{s_1, \ldots, s_n\}$. We were able to find multiple violations for each algorithm, which is highlighted in Fig.~\ref{fig:triangle}. Note that the frequency of violations is a function of the algorithm. In some cases, it does not take a large sampling of triplets to find violations (PLDA), while in other cases, the occurrences are quite rare (Tom-vs-Pete), requiring a much larger evaluation.

\begin{figure}[t]
\centering
\includegraphics [width=1.0\linewidth]{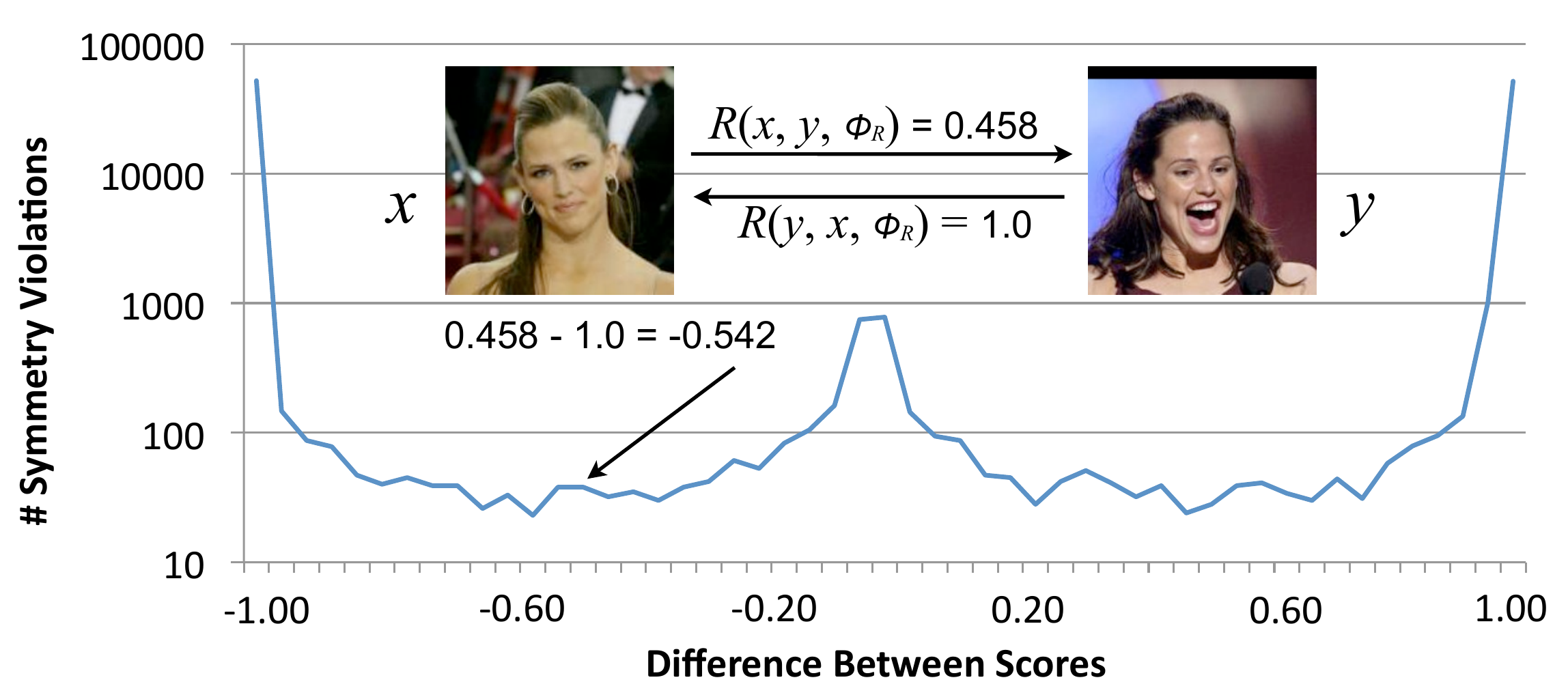}
\caption{Violations in symmetry for the Multi-Attribute Spaces algorithm~\cite{scheirer_cvpr2012} applied over the Image-Restricted Training Protocol of LFW. For each image pair, we calculated the score for image $A$ matching against image $B$, and vice versa. If the scores subtracted from one another do not equal 0, they are considered a violation. Here we see all instances of violation, organized by value of difference. \comment{Multi-Attributes spaces make use of local neighborhoods around \textit{one} particular image in the pair to define a normalization used in the matching process, with no guarantee of symmetry as the neighborhood changes to the other image.}}
\label{fig:lfwsymm}
\end{figure}

Understanding why these violations occur in a seemingly metric scenario is important. Similar to the MkNN algorithm discussed in Sec.~\ref{sec:lfw-meta-analysis}, the Multi-Attribute Spaces algorithm makes use of a local neighborhood of scores around \textit{one} particular image (bounded from below by a parameter $\alpha$, and from above by $\beta$) during a match, in order to build a good model for its normalization~\cite{scheirer_cvpr2012}. Thus, if image $x \neq y$, symmetry is violated in the general case: $\phi_{M}( \alpha_x, \beta_x) \subseteq \{\forall s \in \real : \alpha_x \le s \le \beta_x\} \neq \phi_{M}(\alpha_y, \beta_y) \subseteq\{\forall s \in \real : \alpha_y \le s \le \beta_y\}$. Fig.~\ref{fig:lfwsymm} shows the prevalence of symmetry violations in the Image Restricted Training protocol of LFW. This also means there is no guarantee that the triangle inequality will be satisfied: the local neighborhood considered when matching $(x,y)$ \& $(x,z)$ will differ from $(y,z)$, often resulting in sets of distances that cause a violation. Even under the weaker constraints of quasimetrics and semimetrics, the algorithm still does not satisfy what is necessary to be considered either. Since the Multi-Attribute Spaces algorithm intentionally exploits similarity around single image targets, it is unclear what advantage, if any, would be provided by enforcing the constraints of symmetry and the triangle inequality.

\section{Discussion}

Our meta-analysis, which often required digging very deeply into papers that at first glance had titles and terminology suggesting a metric algorithm, revealed that many such cases are in fact non-metric in their final scoring. Def.~\ref{def:recognition} and a clear specification of where to test for metric properties should help resolve such ambiguity going forward.  In some of the truly metric cases, we cannot actually tell if enforcing metric properties helped or hurt performance. When approaching recognition, it is important that we understand what leads to improvement, and what is tangential.

During the course of this work, we found that some problems and their corresponding solutions do not even have the structural form necessary to be metric -- they compare input features to more complex models.  Similar observations have been made. In~\cite{boiman2008defense} it is proved ``that under the Naive-Bayes assumption, the \textit{optimal} distance to use in image classification is the KL ``Image-to-Class" distance, and not the commonly used ``Image-to-Image" distribution distances.''  Moreover, even for the restricted recognition problem of matched pairs, which at least initially looks as if it is metric, the best performing algorithms have a model for ``matched pairs'' that is non-metric. Metric properties allow some powerful mathematical machinery to be employed and, with effort, any recognition problem's solution can be  ``made'' metric -- the real question is if employing metric constraints improves recognition performance. Our meta-analysis and experimental analysis of top-performing algorithms show violations of symmetry for some and violations of the triangle inequality for others. With so many cases where performance improves as metric conditions are relaxed, we conclude that, in general, \textit{good recognition is non-metric}.

However, this paper should {\em not} be interpreted as suggesting that metrics have no role in computer vision or that metric learning is not useful for recognition. On the contrary, our analysis has shown that metric learning has provided interesting first cut solutions. Furthermore, many good recognition algorithms use local metrics as the core of an overall non-metric algorithm. Learning metrics, at least locally, appears to be an effective way to incorporate various type of constraints. In many cases, the original feature space (Eq.~\ref{eq:feature}) is transformed into another locally normalized/metric feature space, before combining data, yielding a non-metric but effective scoring process.

One observation, which can be exploited in other vision work, is why we believe the problem is inherently non-metric.  General recognition problems need to capture and model the uncertainty in the data and in the class definitions. They must handle local variations in features, in sample density and in labeling.  If, as is true in the general setting, the data is not uniformly sampled with uniform error, good recognition algorithms develop local distance measures in a way that may result in asymmetric measures and/or measures that violate the triangle inequality. Thus, even if one chooses to use local metric learning to help normalize the data, one should also look for models that integrate multiple sources of information (including side-information and other assumptions) and use them to model the regional variations and errors.

A good metric-based recognition algorithm would need to have approximately uniform error. If its ``learning'' could transform an inherently non-uniform biased sampling and errors into a single representation with uniform errors, it would provide a near perfect ``whitening'' filter correcting the per-class biases and errors.  While it is true that in the limit, assuming i.i.d. samples, a metric + nearest neighbor classification has an error rate no more than twice the Bayes error rate, we note that ``in the limit' the infinite i.i.d. sampling requirement is effectively removing any sampling bias and providing uniform error.  Most recognition problems do not have the luxury of i.i.d. sampling nor can they wait for the limit of infinite samples. Thus we believe it is important that computer vision researchers develop robust features and models of uncertainty/error to design more effective recognition algorithms. Finally, we should not stray far from the observations of Tversky, who states that metrics, ``which enhance the interpretability and appeal of spacial representations, cannot always be accepted as valid principles of psychological similarity"~\cite{Tversky_1982}.

{
\bibliographystyle{ieee}
\bibliography{nonmetric_spaces}
}

\end{document}